\documentclass[letterpaper, 10 pt, conference]{ieeeconf}  % Comment this line out if you need a4paper

\IEEEoverridecommandlockouts                              % This command is only needed if 
\usepackage{graphicx} % for pdf, bitmapped graphics files
\usepackage[caption=false,font=footnotesize]{subfig}
\usepackage{times} % assumes new font selection scheme installed
\usepackage{amsmath} % assumes amsmath package installed
\usepackage{amssymb}  % assumes amsmath package installed
\usepackage{url}
\usepackage[breaklinks,hidelinks]{hyperref}
\usepackage{algorithm}
\usepackage{algorithmic}
\usepackage[noadjust]{cite}

\usepackage[T1]{fontenc}

\usepackage{lipsum}

\def\ie{\textit{i.e.,\ }}
\def\eg{\textit{e.g.,\ }}

\newif\ifarxiv
\arxivtrue % uncomment for arxiv

\ifarxiv
\usepackage{tikz}
\fi
\title{\LARGE \bf
CRESSim--MPM: A Material Point Method Library for Surgical Soft Body Simulation with Cutting and Suturing
}

\author{Yafei Ou$^{1}$ and Mahdi Tavakoli$^{1,2}$% <-this % stops a space
\thanks{This research was supported by the Canada Foundation for Innovation (CFI), the Natural Sciences and Engineering Research Council (NSERC) of Canada, the Canadian Institutes of Health Research (CIHR), Alberta Innovates, the China Scholarship Council (CSC), and Alberta Advanced Education. \textit{(Corresponding author: Yafei Ou.)}}% <-this % stops a space
\thanks{$^{1}$Yafei Ou and Mahdi Tavakoli are with the Department of Electrical and Computer Engineering, University of Alberta, Edmonton, Alberta, Canada (e-mail: {\tt\small \{yafei.ou, mahdi.tavakoli\}@ualberta.ca}).}% <-this % stops a space
\thanks{$^{2}$Mahdi Tavakoli is also with the Department of Biomedical Engineering, University of Alberta, Edmonton, Alberta, Canada.}%
}

\begin{document}

\maketitle
\thispagestyle{empty}
\pagestyle{empty}

%=============== arXiv =================
\ifarxiv
% Floating Text Box
\begin{tikzpicture}[remember picture, overlay]
\node [align=left, xshift=10cm, yshift=-0.5cm] at (current page.north west) % Positioning
{
\begin{minipage}{19cm} % Adjust the width of the text box
% \fbox{% You can remove \fbox if you don't want a border
\footnotesize
\textcopyright 2025 IEEE. Personal use of this material is permitted.  Permission from IEEE must be obtained for all other uses, in any current or future media, including reprinting/republishing this material for advertising or promotional purposes, creating new collective works, for resale or redistribution to servers or lists, or reuse of any copyrighted component of this work in other works.
% }
\end{minipage}
};
\end{tikzpicture}
\fi
%=============== arXiv =================

%%%%%%%%%%%%%%%%%%%%%%%%%%%%%%%%%%%%%%%%%%%%%%%%%%%%%%%%%%%%%%%%%%%%%%%%%%%%%%%%
\begin{abstract}
A number of recent studies have focused on developing surgical simulation platforms to train machine learning (ML) agents or models with synthetic data for surgical assistance.
While existing platforms excel at tasks such as rigid body manipulation and soft body deformation, they struggle to simulate more complex soft body behaviors like cutting and suturing.
A key challenge lies in modeling soft body fracture and splitting using the finite-element method (FEM), which is the predominant approach in current platforms.
Additionally, the two-way suture needle/thread contact inside a soft body is further complicated when using FEM.
In this work, we use the material point method (MPM) for such challenging simulations and propose new rigid geometries and soft-rigid contact methods specifically designed for them.
We introduce CRESSim-MPM, a GPU-accelerated MPM library that integrates multiple MPM solvers and incorporates surgical geometries for cutting and suturing, serving as a specialized physics engine for surgical applications.
It is further integrated into Unity, requiring minimal modifications to existing projects for soft body simulation.
We demonstrate the simulator's capabilities in real-time simulation of cutting and suturing on soft tissue and provide an initial performance evaluation of different MPM solvers when simulating varying numbers of particles. The source code is available at \url{https://github.com/yafei-ou/CRESSim-MPM}.
\end{abstract}

%%%%%%%%%%%%%%%%%%%%%%%%%%%%%%%%%%%%%%%%%%%%%%%%%%%%%%%%%%%%%%%%%%%%%%%%%%%%%%%%
\section{INTRODUCTION}
\label{sec:introduction}
Realistic and efficient simulators are playing an increasingly important role in robotics research, not only because they serve as a playground for validating robot control and automation methods, but also because they provide as much synthetic data as possible that can be used for training and testing machine learning (ML) models \cite{tagliabue2020SoftTissue,chiu2021bimanual,bendikas2023learning,kim2024surgical,zargarzadeh2025FromDecision,ou2024LearningAutonomous}, reducing the need of real-world dataset.

While there are numerous high-performance and real-time simulators for other robotics applications, such as autonomous driving (\eg CARLA) and general robot manipulation (\eg Isaac Sim), there are limited resources in terms of surgical robotics simulation.
Although simulating rigid surgical instruments is usually not challenging, the simulation of surgical scenes and operations requires specific simulation capabilities less present in other robotics fields.

The nature of surgery requires simulating diverse objects, involving rigid bodies (\eg surgical instruments and bones), soft bodies (\eg soft tissue), and fluids (\eg blood and smoke).
Additionally, a unique range of manipulation is needed, including cutting, suturing, burning (cauterization), fluid suction, and irrigation, among others.
Simulating and rendering each of these objects and the manipulation types can be a unique research field, making it particularly challenging to develop simulators that meet a wide range of these requirements for surgical robotics research.
As a large group of surgical robotics researchers focus on increasing the level of autonomy in surgeries using ML, the lack of efficient and realistic surgical simulators remains a significant barrier.
\begin{figure}[t]
    \centering
    \includegraphics[width=0.6\columnwidth]{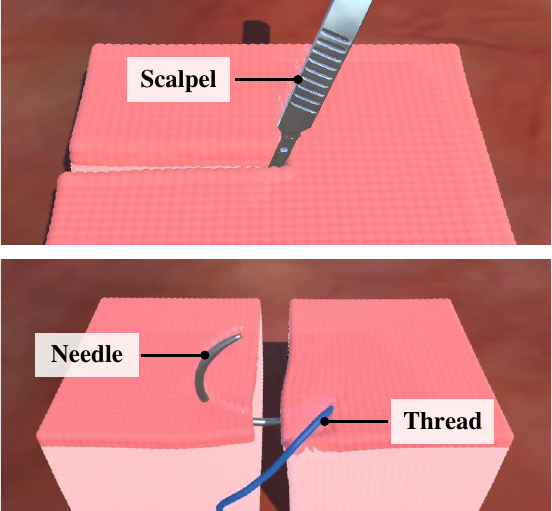}
    \caption{Simulation of soft tissue cutting (upper) and suturing (lower).}
    \label{fig:intro_cover}
\end{figure}

Simulating soft tissue and the corresponding surgical interactions such as cutting and suturing is an extensively explored area. A large amount of previous work utilizes the finite element method (FEM) for simulating soft bodies, including the widely used SOFA Framework \cite{faure2012SOFAMultiModel}. While FEM is accurate for biomedical modeling, its expensive computational cost makes it difficult to achieve efficient and parallel simulation often needed by robotics researchers.
Interactions such as cutting and suturing further require real-time tetrahedron re-meshing algorithms and complex contact constraints when using FEM, making it more computationally expensive.
FEM also suffers from inaccuracies and instabilities under large deformations required in surgical simulation due to the distorted finite elements.
Additionally, while GPUs can accelerate certain FEM computations, achieving full parallel efficiency remains difficult despite ongoing efforts \cite{comas2008EfficientNonlinearFEM}.

The material point method (MPM) \cite{sulsky1995Applicationa} is a promising yet underexplored approach for surgical simulation.
The technique, together with its variations and improvements, has shown wide application in diverse fields including solid mechanics, chemical physics, and computer animation.
MPM utilizes material points (particles) and a background grid for simulating continuum materials, including both solids and fluids.
Despite being a hybrid approach using both Lagrangian particles and an Eulerian grid, it is often viewed as a meshfree method \cite{devaucorbeil2020MaterialPoint} and largely mitigates the disadvantages of the mesh-based FEM.
Although it can be more computationally expensive than FEM if implemented on the CPU, its efficiency can be boosted significantly by GPU acceleration.
It performs well under large deformations and can simulate fracture without re-meshing.
As will be shown in this work, it also enables straightforward rigid coupling for simulating interactions such as cutting and suturing.

Leveraging these strengths, we now present CRESSim-MPM, a GPU-accelerated MPM library tailored for surgical simulation, enabling both soft tissue cutting and suturing within one solver.
Our contributions include:
\begin{itemize}
    \item A CUDA-based MPM library supporting multiple MPM solvers with C-style APIs for efficient integration with external applications and other physics solvers.
    \item Custom geometries and contact models designed to simulate surgical interactions between MPM soft bodies and scalpels, suture needles, and suture threads.
    \item A Unity package that allows easy use of the library and enables two-way interactions between MPM-based soft bodies and Unity’s built-in rigid body physics.
\end{itemize}
Based on these, we demonstrate:
\begin{itemize}
    \item A preliminary evaluation of the simulation speed using different MPM solvers and number of particles.
    \item Real-time soft tissue cutting and suturing simulation within Unity, as shown in Fig.~\ref{fig:intro_cover}.
    % leveraging Unity’s built-in physics engine for rigid body dynamics and rope simulation.
    \item Teleoperation using the Master Tool Manipulator (MTM) from the da Vinci Research Kit (dVRK)~\cite{kazanzides2014OpensourceResearch}.
\end{itemize}
To the best of our knowledge, this is also the first work to simulate soft tissue suturing using the MPM.
Our method couples 1D suturing geometries (\ie an arc-shaped needle with segmented suture thread) with soft tissue dynamics, enabling two-way coupling between MPM soft bodies and 1D rigid bodies.
This formulation also lays the groundwork for incorporating other solvers, such as position-based dynamics (PBD), for future extensions.
The use of MPM further allows for easy future integration of fluids, such as blood and smoke in surgeries, as MPM naturally supports fluid simulation \cite{devaucorbeil2020MaterialPoint}.

CRESSim-MPM is part of the CRESSim project\footnote{\url{https://tbs-ualberta.github.io/CRESSim/}} \cite{ou2024RealisticSurgical,ou2024LearningAutonomous}.

\section{RELATED WORK}
\subsection{Surgical Simulators and Surgical Robotics}
Surgical simulator development, particularly in the area of software-based (virtual) simulators, has mainly been influenced by two streams. The first one originates from the need for surgical training and skills assessment within a simulated environment, with the users being surgeons, residents, and medical researchers.
This category primarily includes virtual simulators and hybrid simulators that are capable of simulating a wide range of surgical training tasks, such as cutting, suturing, and cauterization.
Examples of these are the commercial ones, such as da Vinci SimNow\footnote{\url{https://www.intuitive.com/products-and-services/da-vinci/learning/simnow/}}
% and VirtaMed LaparoS\footnote{\url{https://www.virtamed.com/products-and-solutions/simulators/laparos}},
and the open-source ones such as iMSTK\footnote{\url{https://www.imstk.org/}}.
However, they are less used by robotics researchers due to their proprietary nature, the lack of real-time performance, or difficulty in integrating with existing software, such as the Robot Operating System (ROS).

The second stream arises from the recent trend in surgical robotics research, particularly for surgical automation and autonomy.
Studies including AMBF \cite{munawar2019RealTimeDynamic}, ATAR \cite{enayati2018RoboticAssistanceasNeeded}, and V-Rep Simulator for the dVRK \cite{fontanelli2018VREPSimulator} provide diverse applications for surgical robotics research, such as for benchmarking, synthesizing data, and trial-and-error learning.
Other works such as ORBIT-Surgical \cite{yu2024OrbitSurgicalOpenSimulationa}, Surgical Gym \cite{schmidgall2024SurgicalGym}, FF-SRL \cite{dallalba2024FFSRLHigh}, LapGym \cite{scheikl2023LapGymOpen}, AMBF-RL \cite{varier2022AMBFRLRealtime}, SurRoL \cite{xu2021SurRoLOpensource}, dVRL \cite{richter2020OpenSourcedReinforcement}, and UnityFlexML \cite{tagliabue2020SoftTissue} has specific focus on machine learning applications.
These platforms typically rely on an existing physics engine, such as PhysX (used in Nvidia Omniverse), FleX, SOFA, and Bullet.
Therefore, their simulation capability largely depends on the physics solvers. For instance, Bullet is primarily a rigid body engine and only provides limited FEM soft body support.
Both PhysX 5 and SOFA use FEM for soft body simulation. While PhysX 5 supports FEM, it has yet to achieve realistic interactions such as cutting and suturing. SOFA does support cutting but incurs high computational costs.
These challenges may stem from inherent limitations of FEM itself, as discussed earlier.
% FleX uses PBD, which may be unsuitable for certain surgical scenarios due to its limited accuracy.
A recent study \cite{yang2024efficient} integrates basic MPM support into SurRoL using MLS-MPM but does not address cutting and suturing contact methods.
In contrast, our work focuses on developing a comprehensive engine-level MPM library specifically designed for simulating surgical soft-body manipulations.

\subsection{Real-Time Soft Body Simulation}
The commonly used soft body simulation methods for real-time applications include FEM, MPM, and PBD.
As discussed in Section~\ref{sec:introduction}, FEM has several disadvantages when used for large-scale simulations in robotic research, including computational expenses, the need for real-time re-meshing for topological change, and inaccuracies and instabilities under large deformation.

PBD \cite{muller2007PositionBaseda} and extended PBD (XPBD) \cite{macklin2016XPBDPositionbased} are efficient and numerically stable alternatives, which solve only positional constraints without relying on force integration, making the simulation highly stable.
However, this also means less physically accurate behavior, and the realism of deformations depends on the formulation of constraints rather than the underlying physics.
Although this is acceptable for computer animation, the application in biomedical and robotics research poses additional challenges, as careful parameter tuning is needed to match the simulation with real-world tissue behavior, as in \cite{liu2021RealtoSimRegistration}.
Additionally, forming positional constraints can sometimes be challenging for certain types of materials, such as viscoelastic or elastoplastic models, which are common in biomedical applications.

MPM overcomes many disadvantages of FEM while preserving accuracy and versatility for simulating a wide range of material behaviors.
MPM is considered within the particle-in-cell (PIC) methods family, where particles carry physical quantities and a background grid is used for computations.
The initial PIC method \cite{francish1964ParticleincellComputing} was developed for fluid mechanics, with later improvements by the fluid implicit particle method (FLIP) \cite{brackbill1986FLIPMethod}.
MPM extends the methods to simulating solids and multiphase materials, including elastic and hyperelastic materials, granular materials (\eg snow--used by Disney in their film \textit{Frozen} \cite{stomakhin2013MaterialPoint}--and sand), and elastoplastic materials such as biological tissues.
In addition, since MPM does not explicitly use a mesh, topological changes are not a concern, making it easier to simulate cutting and fracture.
The MPM and the PIC methods are related and have numerous variations such as generalized interpolation material point (GIMP) \cite{bardenhagen1970GeneralizedInterpolation}, convected particle domain interpolation (CPDI) \cite{sadeghirad2011ConvectedParticle}, affine particle-in-cell (APIC) \cite{jiang2015AffineParticleincell}, moving least squares MPM (MLS-MPM) \cite{hu2018MovingLeast}, and more recently position-based MPM (PB-MPM) \cite{lewin2024PositionBased}.
However, MPM is less investigated and adopted for surgical simulation, possibly due to the lack of a high-performance library specifically built for surgical applications.

\section{MATERIAL POINT METHOD}
\label{sec:mpm}
\subsection{Standard MPM}
MPM is an approach without explicit consideration of a computational mesh as in FEM. Instead, a background Eulerian (\ie fixed) grid is used for force and momentum computations, whereas a continuum body is discretized into Lagrangian material points (\ie moving particles) to store local material data, such as mass, velocity, stress tensor, and deformation gradient.

\begin{figure}[ht]
    \centering
    \includegraphics[width=0.75\columnwidth]{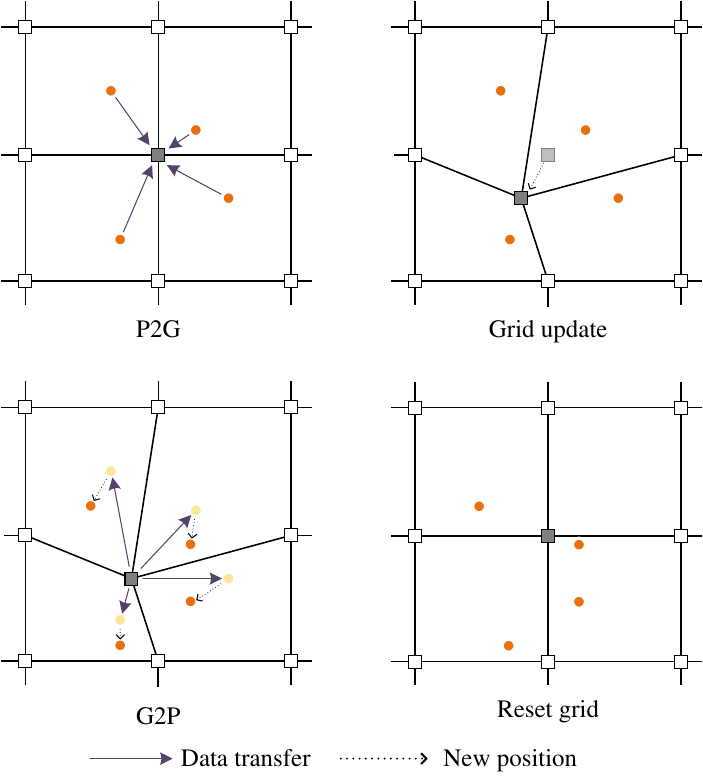}
    \caption{Overview of the standard MPM formulation, adapted from \cite{devaucorbeil2020MaterialPoint}.}
    \label{fig:p2g2p}
\end{figure}

The standard formulation of the MPM follows 4 steps: \textbf{(1) Particle to grid (P2G)}: Particle properties are transferred and distributed to the background grid using an interpolation shape function. \textbf{(2) Grid update}: the momentum of each grid node is updated using numerical integration. \textbf{(3) Grid to particle (G2P)}: The updated node momenta are interpolated back to the particles. Particle properties such as the deformation gradient and stress are also updated. \textbf{(4) Reset grid}: clear grid node data to prepare for the next step. This is illustrated in Fig.~\ref{fig:p2g2p}.
As an example, consider a Neo-Hookean hyperelastic model using a PIC-style particle velocity update. At time $t$, particle data is transferred to the grid using
\begin{equation}
\label{eqn:mls_mpm_p2g_mass}
    m_{I}^t = \sum_{i} \phi(\mathbf{x}_i^t) m_i, \;
    (m \mathbf{v})_I^t = \sum_{i} \phi(\mathbf{x}_i^t) (m \mathbf{v})_i^t.
\end{equation}
Here, we use the subscript $i$ to denote the particle indices and $I$ for grid indices. $\mathbf{x}$ is the position, $m$ is the mass, $\mathbf{v}$ is the velocity, and $\phi$ is the shape function.
As the shape function uses a basis function $N(x)$ with finite support, only grid nodes within this region are affected by a specific particle. With a quadratic B-spline basis, a particle can transfer data to 3 neighboring nodes per dimension.
Force is transferred from the particle to the grid using
\begin{equation}
    \mathbf{f}_I^{int,\,t} = -\sum_i V_i^t \boldsymbol{\sigma}_i^t \nabla \phi(\mathbf{x}_i^t),
    \; \mathbf{f}_I^{ext,\,t} = \sum_{i} \phi(\mathbf{x}_i^t) m_i \mathbf{b}(\mathbf{x}_i^t),
\end{equation}
where $\mathbf{f}_I^{int}$ is the internal force and $\mathbf{f}_I^{ext}$ is the external force. $V_i$ is the particle volume, $\boldsymbol{\sigma}_i$ is the stress tensor, $\mathbf{b}$ is the body force. $\mathbf{b} = \mathbf{g}$ when only gravity is applied. $\nabla \phi$ is the gradient of the shape function.

Next, the grid is updated for each node with
\begin{equation}
    (m \mathbf{v})_I^{t+\Delta t} =  (m \mathbf{v})_I^t + \mathbf{f}_I^t \Delta t,
    \;    \mathbf{v}_I^{t+\Delta t} = (m \mathbf{v})_I^{t+\Delta t} / m_I^t,
    \label{eqn:mls_mpm_grid}
\end{equation}
where $\mathbf{f}_I = \mathbf{f}_I^{int} + \mathbf{f}_I^{ext}$ is the total force. Boundary conditions (\eg fixed nodes) can be applied by altering the momentum.

Updated nodal velocities are then transferred back to particles for integration, with
\begin{equation}
    \mathbf{v}_i^{t+\Delta t} = \sum_I \phi(\mathbf{x}_i^t) \mathbf{v}_I^{t+\Delta t}, \;
    \mathbf{x}_i^{t+\Delta t} = \mathbf{x}_i^{t} + \mathbf{v}_i^{t+\Delta t} \Delta t. \label{eqn:mls_mpm_g2p_integration}
\end{equation}
Finally, particle velocity gradient $\mathbf{L}$, deformation gradient $\mathbf{F}$, volume, and stress are updated using
\begin{align}
    \mathbf{L}_i^{t+\Delta t} &= \sum_I \nabla \phi(\mathbf{x}_i^t) \mathbf{v}_I^{t+\Delta t}, \\
    \mathbf{F}_i^{t+\Delta t} &= (\mathbf{I} + \mathbf{L}_i^{t+\Delta t} \Delta t) \mathbf{F}_i^{t}, \\
    V_i^{t+\Delta t} &= J V_i^t, \; \text{with } J = \det (\mathbf{F}_i^{t+\Delta t}), \\
    \boldsymbol{\sigma}_i^{t+\Delta t} &= \frac{1}{J} \left[ \mu (\mathbf{F} \mathbf{F}^T - \mathbf{I}) + \lambda \ln(J)\mathbf{I} \right]. \label{eqn:neo_hookean}
\end{align}
$\mathbf{I}$ denotes the identity matrix. Equation (\ref{eqn:neo_hookean}) is specific to the Neo-Hookean model for calculating the Cauchy stress, where $\mu$ and $\lambda$ are the Lam\'e parameters.
Some implementations use the Kirchhoff stress instead.

One benefit of MPM is that contact between multiple soft bodies is inherently handled, as each body is represented by a cluster of particles interacting through the shared background grid. This eliminates the need for explicit collision detection.

Since more advanced algorithms are now commonly used, standard MPM is implemented in this work only as a reference with limited feature support.

\subsection{Moving Least Squares MPM}
The moving least squares MPM (MLS-MPM) \cite{hu2018MovingLeast} uses APIC-style transfers and unifies it with force computation by using a moving least squares shape function.
This leads to modifications of the velocity and force transfers in the standard MPM formulation. For the internal force,
\begin{equation}
    \mathbf{f}_I^{int,\,t} = -\sum_i V_i^t M_i^{-1} \boldsymbol{\sigma}_i^t N_I(\mathbf{x}_i^t) (\mathbf{x}_I - \mathbf{x}_i^t),
\end{equation}
where $N_I(x)$ is the same shape basis function used in standard MPM, centered at node $I$. $M_i = \frac{1}{4} {\Delta x}^2$ for quadratic basis and grid cell size $\Delta x$.
APIC affine momentum $\mathbf{C}_i^t$ and force transfer is then fused in P2G, with
\begin{equation}
\label{eqn:mls_mpm_p2g_momentum_force}
    (m \mathbf{v})_I^t = \sum_{i} N_I(\mathbf{x}_i^t) \left(m_i \mathbf{C}_i^t - \Delta t V_i^t M_i^{-1} \boldsymbol{\sigma}_i^t \right)  (\mathbf{x}_I - \mathbf{x}_i^t),
\end{equation}
where $\mathbf{C}_i^t$ can be updated in the APIC way during G2P using
\begin{equation}
\label{eqn:mls_mpm_g2p_momentum}
    \mathbf{B}_i^{t+\Delta t} = \sum_{i} N_I(\mathbf{x}_i^t) v_i^{t + \Delta t} (\mathbf{x}_I - \mathbf{x}_i^t)^T, \;
    \mathbf{C}_i^{t+\Delta t} = \mathbf{B}_i^t \mathbf{D}_i^t.
\end{equation}
For quadratic basis, $\mathbf{D}_i^t \equiv \frac{1}{4}{\Delta x}^2\mathbf{I}$, resulting in a scaling on $\mathbf{B}_i^t$.
The deformation gradient is then updated with
\begin{equation}
\label{eqn:mls_mpm_g2p_f}
    \mathbf{F}_i^{t+\Delta t} = (\mathbf{I} + \mathbf{C}_i^{t+\Delta t} \Delta t) \mathbf{F}_i^{t}.
\end{equation}
Algorithm~\ref{alg:mls_mpm} shows the structure of APIC-style MLS-MPM.

\begin{algorithm}[H]
\caption{APIC-Style MLS-MPM Simulation Step}
\label{alg:mls_mpm}
\begin{algorithmic}[1]
\FOR{each particle $i$}
    \STATE P2G mass transfer using the first of (\ref{eqn:mls_mpm_p2g_mass}).
    \STATE P2G fused momentum-force transfer using (\ref{eqn:mls_mpm_p2g_momentum_force}).
\ENDFOR
\FOR{each grid node $I$}
    \STATE Update nodal velocity using (\ref{eqn:mls_mpm_grid}) with external force.
    \STATE Modify nodal velocity based on contact and boundary conditions.
\ENDFOR
\FOR{each particle $i$}
    \STATE Affine momentum transfer using (\ref{eqn:mls_mpm_g2p_momentum}).
    \STATE Update particle positions using (\ref{eqn:mls_mpm_g2p_integration}).
    \STATE Update deformation gradient using (\ref{eqn:mls_mpm_g2p_f}).
    \STATE Compute stress using the constitutive law.
\ENDFOR
\STATE \textbf{Reset grid:} Clear grid momentum $(m \mathbf{v})_I$.
\STATE $t \gets t + \Delta t$
\end{algorithmic}
\end{algorithm}

\subsection{Position-Based MPM}
PB-MPM, proposed in 2024 \cite{lewin2024PositionBased}, is a new method that formulates MPM as a constraint-solving problem similar to PBD, allowing larger integration steps and better numerical stability.
The modifications needed to convert any MPM algorithm to PB-MPM are minimal: at each simulation step, multiple P2G-G2P cycles are conducted to iteratively solve a candidate particle velocity gradient.
In P2G, instead of calculating the stress, a material constraint is solved for all particles independently.
For example, for an elastic material with MLS-MPM formulation, the co-rotational constraint-solving step includes
\begin{align}
    \mathbf{F}_i^{t+\Delta t, *} &\gets (\mathbf{I} + \mathbf{C}_i^{t+\Delta t} \Delta t) \mathbf{F}_i^{t}, \\
    \mathbf{R}, \mathbf{U} &= \text{PolarDecomp}( \mathbf{F}_i^{t+\Delta t, *}), \\
    \mathbf{F}_i^{t+\Delta t, *} &\gets \beta \mathbf{R} + (1-\beta) \frac{\mathbf{F}_i^{t+\Delta t, *}}{\det(\mathbf{F}_i^{t+\Delta t, *})}, \\
    % \mathbf{C}_i^{t+\Delta t} &\gets \mathbf{C}_i^{t+\Delta t} + \alpha \left( \mathbf{F}_i^{t+\Delta t, *} (\mathbf{F}_i^{t})^{-1} - \mathbf{I} - \mathbf{C}_i^{t+\Delta t} \right),
    \mathbf{C}_i^{t+\Delta t} &\gets \frac{1}{\Delta t} \left( \mathbf{F}_i^{t+\Delta t, *} (\mathbf{F}_i^{t})^{-1} - \mathbf{I} \right),
\end{align}
which formulates a Jacobi-style constraint projection solver.

PB-MPM integrates the benefits of both PBD and MPM but also inherits certain drawbacks from PBD, such as reliance on non-physical parameters and stiffness that depends on the number of iterations.
Therefore, its application should be considered based on specific use cases.

\subsection{Computational Speed Comparison}
CRESSim-MPM currently implements the aforementioned three types of MPM solvers in 3D for users to select from.
More solver algorithms can be added in a modular manner.

\begin{figure}[ht]
    \centering
    \includegraphics[width=0.7\columnwidth]{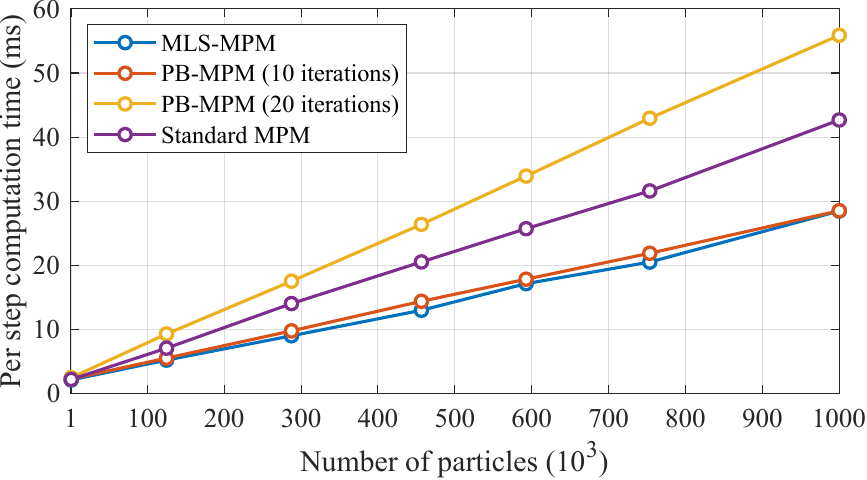}
    \caption{Simulation speed with different numbers of particles.}
    \label{fig:performance}
\end{figure}

We provide an initial performance evaluation when simulating a cubic object falling under gravity with different numbers of particles for the implemented solvers on Nvidia RTX 4090 (compute capability 8.9), with CUDA Toolkit version 12.4.1.
For all solvers, each step advances the simulation by $0.02$ second.
For MLS-MPM and standard MPM, each step contains 10 integration sub-steps with $\Delta t = 0.002$.
For PB-MPM, each step uses $\Delta t = 0.02$ with 10 or 20 iterations.
The grid size is fixed in all situations with $175,616$ nodes.
The results are shown in Fig.~\ref{fig:performance}.
While comparing with existing implementations is reasonable, this work does not aim to outperform existing code in computational speed, as our main focus is on including multiple MPM solver options within one library.
% It is also worth noting that we were unable to find any existing 3D implementation of PB-MPM, likely due to its relatively recent proposal.
Further code optimization and performance benchmarking are planned as future work.

\section{RIGID GEOMETRIES AND CONTACT MODELS FOR SURGICAL SIMULATION}
\subsection{Soft-Rigid Coupling}
\begin{figure}[t]
    \centering
    \includegraphics[width=0.7\columnwidth]{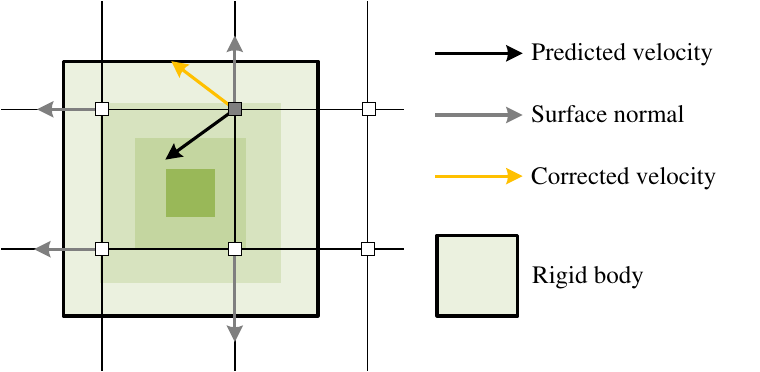}
    \caption{Grid level collision with SDF. Nodal velocities are modified if they are inside a rigid body.}
    \label{fig:rigid_collision}
\end{figure}
Two-way coupling between MPM soft bodies and externally-solved rigid bodies can be achieved at the grid level \cite{stomakhin2013MaterialPoint,hu2018MovingLeast}.
The most straightforward approach is to use a signed distance field (SDF) for each rigid body and modify the predicted nodal velocity based on the contact model if a grid node is inside the rigid geometry.
This is similar to a predictor-corrector method \cite{devaucorbeil2020MaterialPoint}, as shown in Fig.~\ref{fig:rigid_collision}. % cite: 25 years;
We set the nodal velocity to zero in the normal direction, and apply frictional correction to the tangential component, unless the node is separating from the rigid body.
Mathematically, the relative velocity between a node and a rigid body is $\mathbf{v}_{rel} = \mathbf{v}_I - \mathbf{v}_{r}$, with $\mathbf{v}_{r}$ being the rigid body's velocity. Given a rigid surface normal $\mathbf{n}$, $\mathbf{v}_{rel}$ is decomposed into the tangent component $\mathbf{v}_{tg}$ and the normal part $v_n$:
\begin{equation}
    \mathbf{v}_{rel} = \mathbf{v}_{tg} + \mathbf{n} v_n.
\end{equation}
If the node is separating from the rigid body, the velocity is not changed, \ie $\mathbf{v}'_{rel} = \mathbf{v}_{rel}$. Otherwise,
\begin{equation}
\label{eqn:v_correction}
    \mathbf{v}'_{rel} = c_d \max \left(1 - \frac{\mu_k v_n}{\lVert\mathbf{v}_{tg}\rVert},  0 \right) \mathbf{v}_{tg},
\end{equation}
where $\mu_k$ is the kinetic friction coefficient and $c_d$ applies a linear drag force on the velocity, which is useful for sticky effects such as tissue sticking to the scalpel and needle.

\begin{figure}[ht]
    \centering
    \includegraphics[width=0.5\columnwidth]{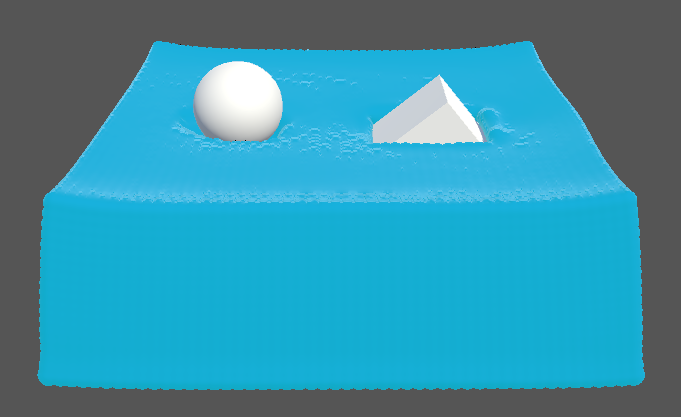}
    \caption{Soft body (blue) coupling with rigid bodies (white sphere and cube) of different masses.}
    \label{fig:demo_rigid_coupling}
\end{figure}

The velocity change $\mathbf{v}'_{rel} - \mathbf{v}_{rel}$ can then be recorded and used as a momentum impulse applied back to the rigid body, yielding a two-way force coupling.
Additional particle correction is performed to push out the remaining particles inside the rigid geometry, similar to~\cite{stomakhin2013MaterialPoint}.
The SDF-based approach works well for MPM-rigid collisions with regular geometries, as shown in Fig.~\ref{fig:demo_rigid_coupling}.
% This work uses the winding number when calculating triangle mesh SDFs.

\subsection{Geometries and Contact Method for Cutting}

For cutting, authors of \cite{hu2018MovingLeast} propose using a colored distance field (CDF).
The CDF stores points that contain an unsigned distance and a \textit{color} information, which encodes a set of nearby surfaces and the side it is located at.
If a particle and a node are located on different sides of a surface, the node is considered \textit{incompatible}, and no P2G and G2P transfer occurs.
This forms a discontinuity that blocks energy transfer at the cutting surface.

While this approach allows cutting with a random surface geometry, it requires the additional storage of nearby geometries and the use of rigid particles.
This work simplifies the approach and uses a regular SDF representation for all rigid geometries.
For a cutting geometry, the sign of the SDF value represents the side of the point relative to the nearest surface.
This eliminates the need to store additional data for cutting geometries and particles and unifies the grid-level contact resolution by simply using the absolute value of the SDF for contact.
However, as the absolute value is always positive, a fattened collision region is needed during contact resolution, as opposed to using CDF.

\begin{figure}[t]
    \centering
    \subfloat[]{%
    \includegraphics[width=0.3\columnwidth]{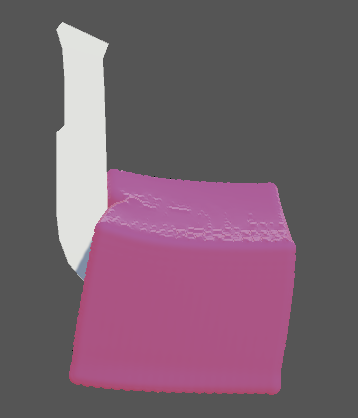}%
    }\hfil
    \subfloat[]{%
    \includegraphics[width=0.3\columnwidth]{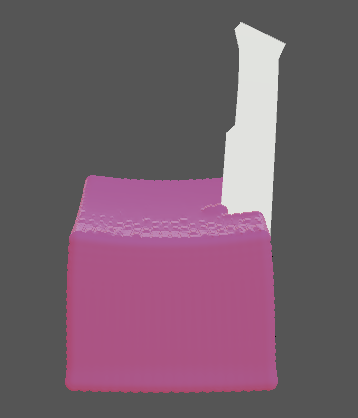}%
    }
    \caption{Contact between a soft body and a slicer geometry on (a) the spine side and (b) the edge side.}
    \label{fig:demo_cutting}
\end{figure}

\begin{figure}[t]
    \centering
    \includegraphics[width=0.7\columnwidth]{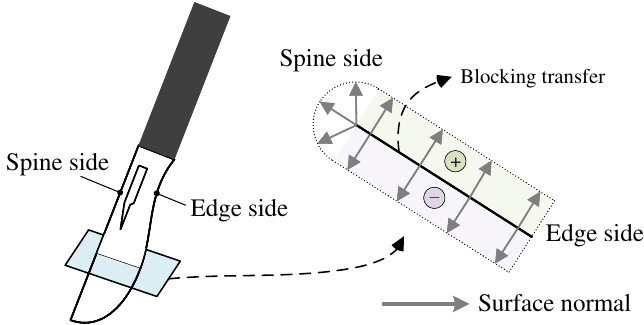}
    \caption{SDF for a slicer geometry. The signs of the distance on both sides are different.}
    \label{fig:slicer}
\end{figure}

Additionally, since a knife or scalpel can cut with only the edge side instead of the spine (blunt) side, it is inappropriate to have a geometry that cuts in all directions.
Considering this, we apply a circular-shaped distance field to the spine side and use sticky contact (zero relative velocity) if a node is in this region.
When the spine side touches an MPM soft body, it generates contact similar to a cylinder colliding with the soft body, as shown in Fig.~\ref{fig:demo_cutting}.
The SDF of a scalpel's cross-section is illustrated in Fig.~\ref{fig:slicer}.
The geometry is named \textit{slicer} in our code implementation, which includes \texttt{QuadSlicer} and \texttt{TriangleMeshSlicer}.

\subsection{Geometries and Contact Method for Suturing}
\begin{figure}[ht]
    \centering
    \includegraphics[width=0.55\columnwidth]{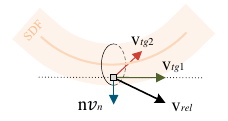}
    \caption{Nodal velocity decomposition for an arc geometry. The velocity is projected on the tangent direction of the arc and then corrected with friction and drag force.}
    \label{fig:1d_geom}
\end{figure}

\begin{figure}[ht]
    \centering
    \subfloat[]{%
    \includegraphics[width=0.9\columnwidth]{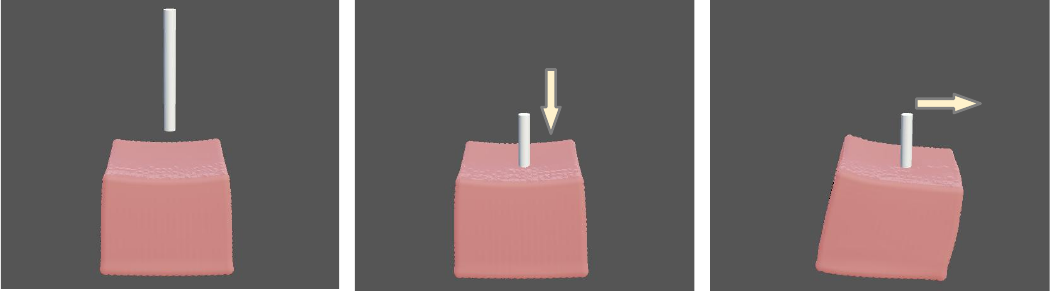}%
    }\\
    \subfloat[]{%
    \includegraphics[width=0.9\columnwidth]{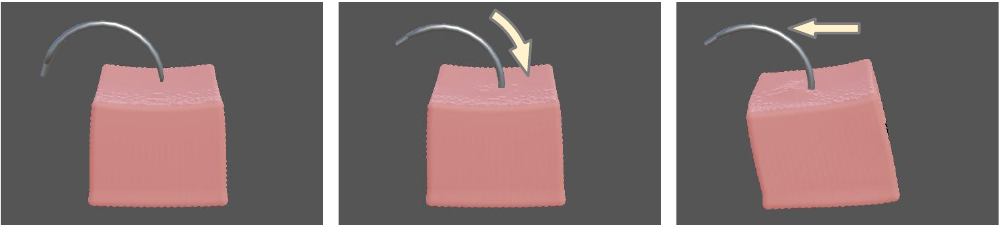}%
    }
    \caption{1D geometries can move freely along their tangent directions in a soft body, but deform the soft body when moving in the lateral direction.}
    \label{fig:1d_geometries}
\end{figure}

In this work, we assume that the suture thread can be discretized into connected straight-line segments.
To achieve suturing, we introduce the arc and the line segment geometries, which can model an arc-shaped suture needle and a discrete thread segment.
They are both 1D geometries with SDFs.
The signed distance is calculated analytically from a given query point to its nearest projected point on the geometry, which is always positive.
The surface normal $\mathbf{n}$ is the normalized vector from the projected point to the query point.
Nodal velocities inside the SDF region are corrected differently than for regular geometries.
The relative velocity is decomposed into three parts: one along the arc or line tangent $\mathbf{v}_{tg1}$, one that lies on the cross-section of the geometry $\mathbf{v}_{tg2}$, and one along the surface normal $\mathbf{n} v_n$, as shown in Fig.~\ref{fig:1d_geom}, and only the tangent component is kept.
Mathematically, the relative velocity $\mathbf{v}_{rel}$ is decomposed to
\begin{equation}
    \mathbf{v}_{rel} = \mathbf{v}_{tg1} + \mathbf{v}_{tg2} + \mathbf{n} v_n,
\end{equation}
where $\mathbf{v}_{tg1}$ is the true tangent velocity along the arc or line. Nodal velocity correction is applied by replacing $\mathbf{v}_{tg}$ with $\mathbf{v}_{tg1}$ in (\ref{eqn:v_correction}).
This ensures that the rigid geometry can move inside a soft body if the motion is along the geometry (\eg inserting a needle) while preventing lateral motion, as shown in Fig.~\ref{fig:1d_geometries}.
Geometries for suturing include \texttt{Arc} and \texttt{ConnectedLineSegments}, where the latter is a generalized geometry of a line segment.

Various external methods, such as the kinematic chain of rigid links with joints \cite{brown2004RealtimeKnottying}, mass-spring model \cite{leduc2003ModelingSuturing}, and PBD~\cite{munawar2022OpenSimulationa}, can be used to simulate the suture thread itself, as long as the coupling momentum from the soft body can be applied back to each small segment.

\section{IMPLEMENTATION}
\subsection{Object Model and Simulation Lifecycle}
\begin{figure}[ht]
    \centering
    \includegraphics[width=0.65\columnwidth]{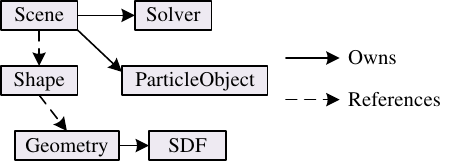}
    \caption{CRESSim-MPM object model.}
    \label{fig:object_model}
\end{figure}

The library follows an object model similar to the one in PhysX, such that integration with other PhysX-based software is straightforward, as shown in Fig.~\ref{fig:object_model}.
\texttt{Scene} contains all simulation objects and provides the method \texttt{advance(float dt)} for starting to advance the simulation by a time step \texttt{dt}, and the method \texttt{fetchResults()} retrieves the results of the simulation, blocking the thread until the results are available.
This is similar to PhysX's simulation lifecycle which splits the simulation into two steps, allowing asynchronous tasks to run while the solver advances.

\subsection{Implementation Details and Usage}
The library is implemented in {CUDA C++} and is dependent only on the standard C++ library and the CUDA runtime.
3 types of \texttt{Solver} discussed in Section~\ref{sec:mpm} are currently included.
\texttt{ParticleObject} represents an object of material points simulated by MPM, and \texttt{Shape} is a rigid body.
\texttt{SimulationFactory} is used to create all objects.
C++ Users should only include the header \texttt{simulation\_factory.h}.
A backend C-style API engine is also provided for integration into existing software.

While we also provide a CPU version for each solver, they are not optimized for multi-core parallelism or single instruction, multiple data (SIMD). Users who require simulations that fully utilize the CPU should refer to alternative software, such as Taichi-MPM\footnote{\url{https://github.com/yuanming-hu/taichi_mpm}} for MLS-MPM, and MPM-GIMP\footnote{\url{https://sourceforge.net/p/mpmgimp}} for standard MPM and GIMP.

\subsection{Unity Integration}
\label{sec:unity_integration}
\begin{figure}[ht]
    \centering
    \subfloat[]{%
    \includegraphics[height=0.32\columnwidth]{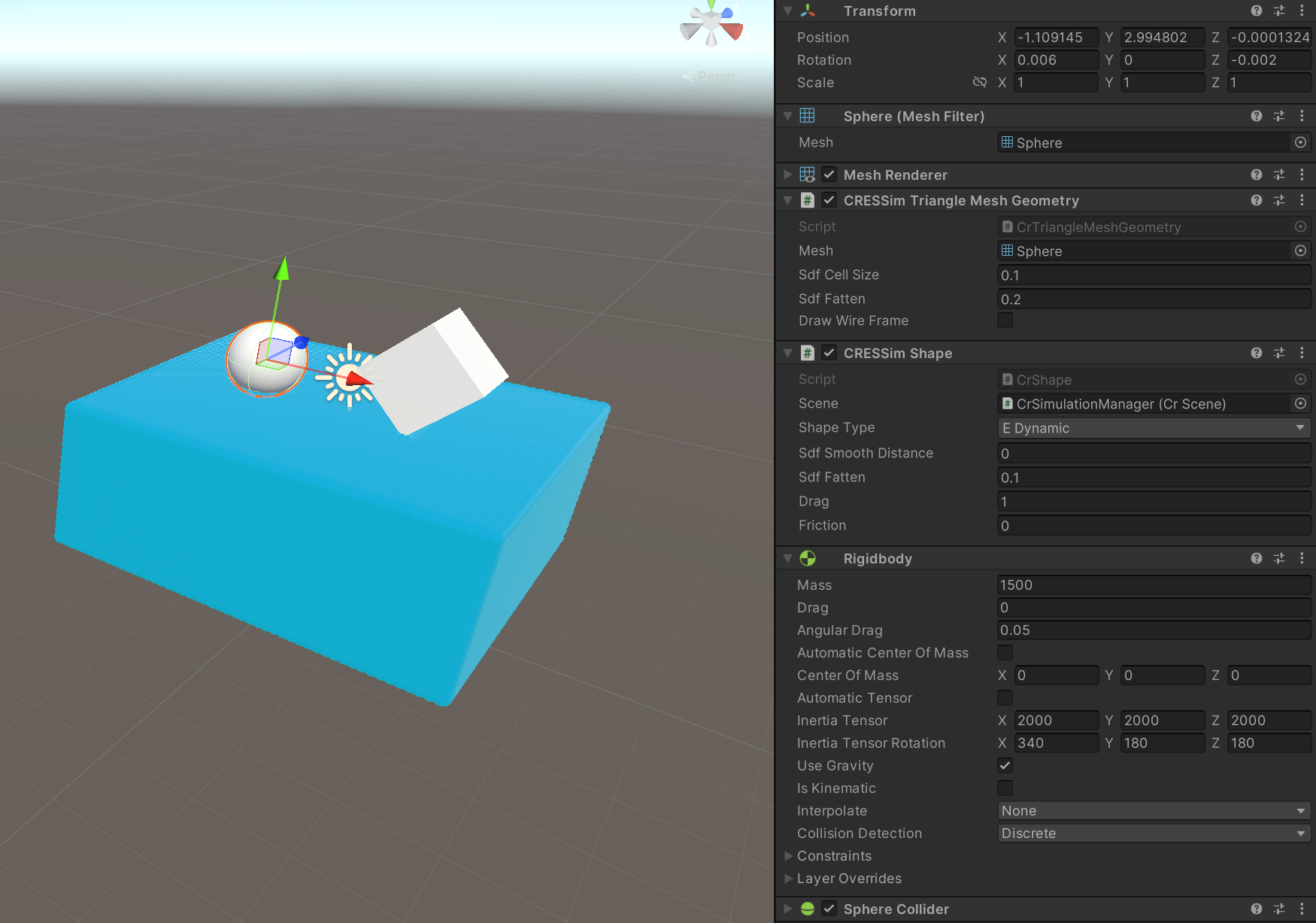}%
    \label{fig:unity_integation}
    }\hfil
    \subfloat[]{%
    \includegraphics[height=0.32\columnwidth]{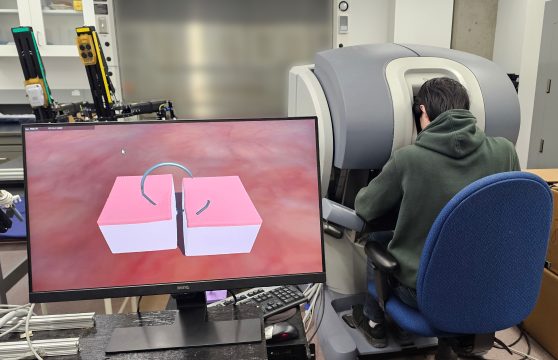}
    \label{fig:teleop}
    }
    \caption{Usage of the library: (a) Unity integration; (b) Teleoperating the suturing scene using the dVRK MTM.}
\end{figure}

Integration into Unity is straightforward using the C-style APIs.
Only a few modifications are needed to allow the coupling between the MPM soft body and rigid bodies simulated in Unity's default engine.
For each rigid body in Unity, a corresponding \texttt{Shape} is created for two-way coupling with the soft body.
During each step, the rigid body is simulated first with Unity, after which the \texttt{Shape}'s velocity is updated.
The MPM engine is then advanced, recording a total momentum change of the contact grids with the rigid body.
This is then used to apply a momentum impulse on the rigid body back in Unity.
A Unity plugin is provided as part of CRESSim-MPM, as shown in Fig.~\ref{fig:unity_integation}.

\section{SURGICAL SIMULATION EXAMPLES}
This section includes screenshots from two example scenes representing surgical cutting and suturing. While the visuals and rendering are not highly realistic, the examples are primarily designed to showcase the physics capabilities.
For demonstration purposes, we use two different solvers, although either one works in both scenarios.
Additional examples are provided in the supplementary video.

\subsection{Cutting (PB-MPM)}
\begin{figure}[ht]
    \centering
    \includegraphics[width=0.85\columnwidth]{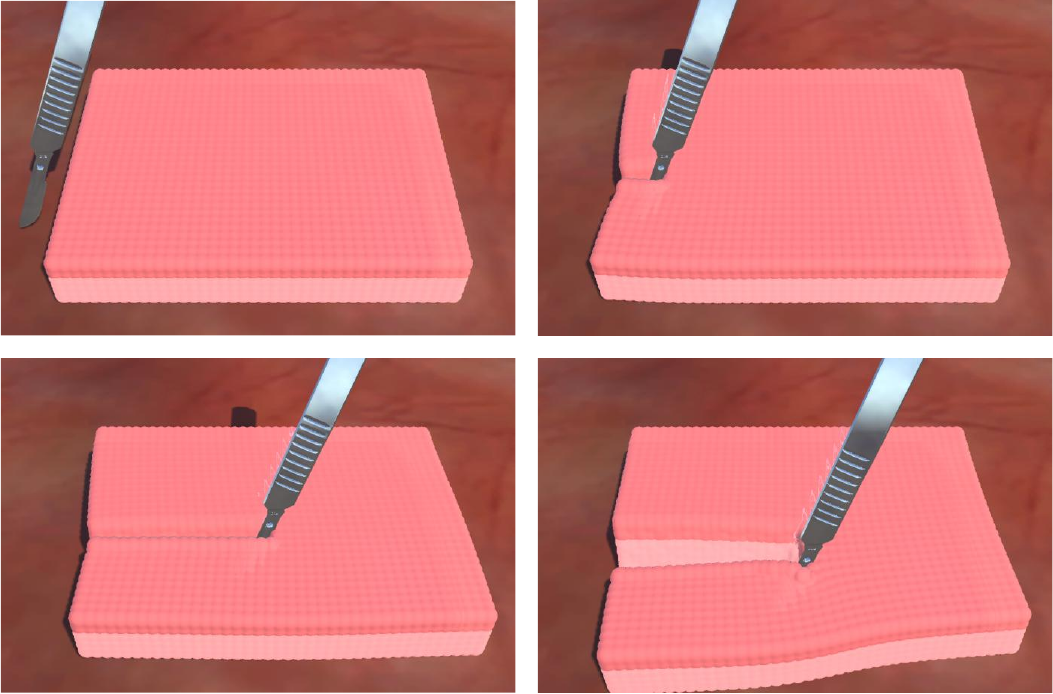}
    \caption{Soft tissue cutting with a scalpel moving right then down.}
    \label{fig:example_cutting}
\end{figure}

This example demonstrates that we can simulate soft tissue cutting using a scalpel.
The scalpel moves right to cut the tissue first and then down to split and reveal both sides of the cut tissue.
The screenshots are shown in Fig.~\ref{fig:example_cutting}.

\begin{figure}[t]
    \centering
    \includegraphics[width=0.85\columnwidth]{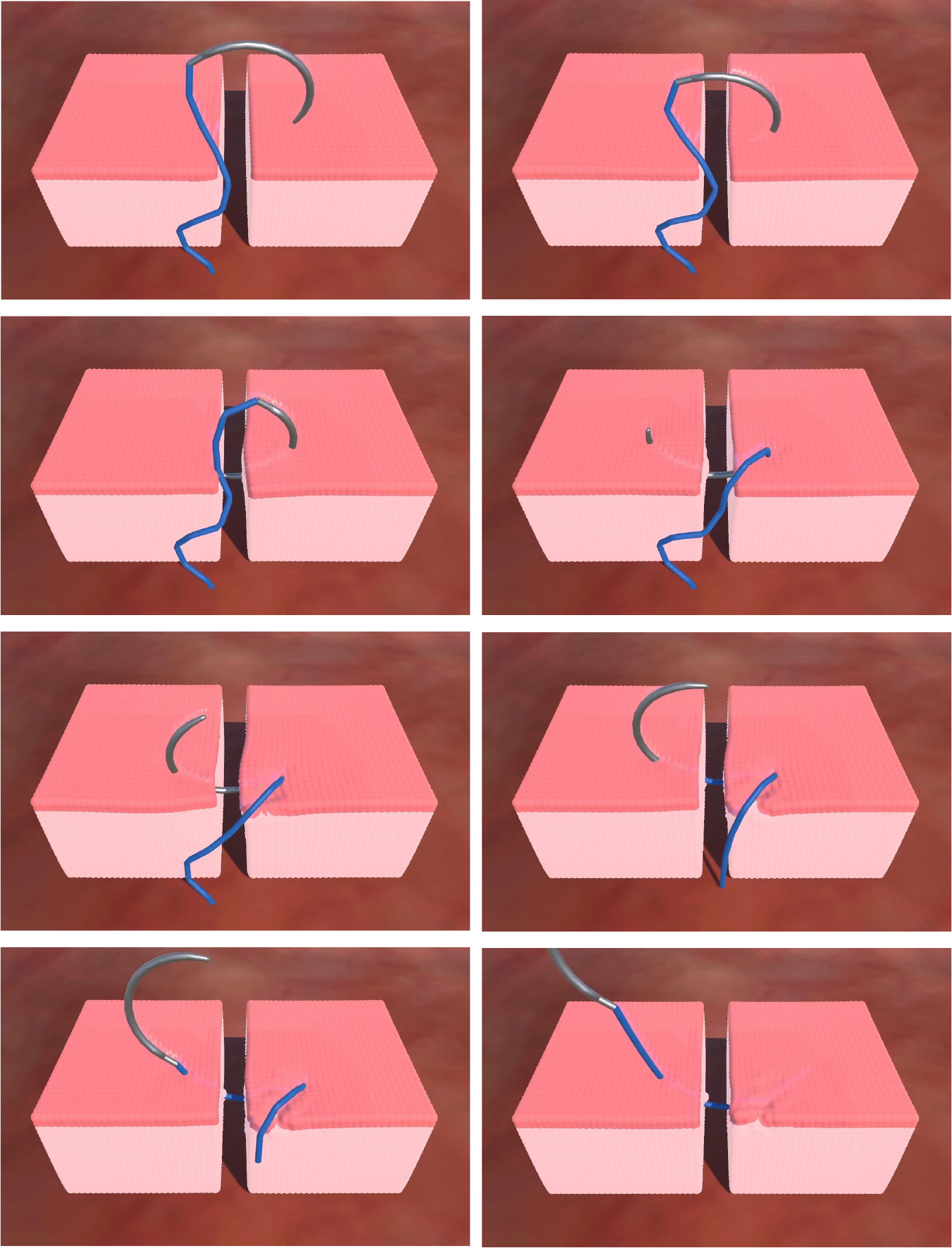}%
    \caption{Suturing between two soft body blocks.}
    \label{fig:example_suturing}
\end{figure}

\subsection{Suturing (MLS-MPM)}
This example demonstrates the capability of simulating soft tissue suturing using an arc-shaped needle.
For the sake of simplicity, we use a kinematic chain of rigid links with spherical joints to represent a suture thread, which is implemented using Unity's default physics engine.
This is a traditional approach that does not provide an adequate level of realism \cite{brown2004RealtimeKnottying,phellan2021RealtimeBiomechanics}. Nevertheless, the focus of this work is not the thread simulation itself.
Integration of more advanced approaches, such as PBD-based threads \cite{munawar2022OpenSimulationa}, is straightforward since contact force coupling can be achieved in a similar manner. % cite AMBF suturing work
Screenshots of this example are shown in Fig.~\ref{fig:example_suturing}.
This example can be teleoperated using the MTM robot from the dVRK, as shown in Fig.~\ref{fig:teleop}.

\section{LIMITATIONS AND FUTURE WORK}
We are aware of several areas where the library can be enhanced, and we're actively working on improving these before an official release.
These include (1) code optimization, particularly through increased use of fused multiply-add (FMA) operations, shared memory for rigid shapes, and improved thread block scheduling; (2) implementation of additional material models, such as elastoplastic materials and fluids, to simulate specific tissue types, blood, and smoke; (3) integration of other MPM and PIC solvers, including APIC and FLIP, as well as PBD solvers for fascia and suture thread simulation; and (4) improved rendering of particle objects.

There is also a lack of performance comparison with existing general-purpose MPM implementations, such as Taichi-MPM which does not support features such as suturing and only uses MLS-MPM.
This is left as a future work after code optimization.
We also recognize that relying on CUDA restricts broader compatibility with computing devices beyond Nvidia GPUs.
Extending to other platforms can require a significant amount of future work, although porting this work to AMD devices should be straightforward using ROCm/HIP.

\section{CONCLUSION}
In this work, we introduced CRESSim-MPM, a new MPM library designed for surgical simulation that implements multiple solvers and surgery-specific contact models for cutting and suturing.
By using it as a separate physics module, integration with existing surgical simulation platforms is straightforward, and we further provide a Unity plugin to allow direct integration into Unity.
With the capability of simulating cutting and suturing on soft tissue efficiently on the GPU, this work enables more realistic and efficient surgical simulations that can be potentially used in various applications, including soft tissue modeling, surgical skills training, and surgical robot learning.

% \addtolength{\textheight}{-2cm} 

\bibliographystyle{IEEEtran}
\bibliography{references.bib}

\end{document}